\documentclass{article}
\usepackage{ijcai19}

\author{Ganbin Zhou$^{1,2,3}$ \and Ping Luo$^{1,2}$ \and Jingwu Chen$^{1,2}$ \and Fen Lin$^{3}$\and Leyu Lin$^{3}$\and Qing He$^{1,2}$ \\
	\affiliations
	$^{1}$ Key Lab of Intelligent Information Processing of Chinese Academy of Sciences (CAS), \\Institute of Computing Technology, CAS, Beijing 100190, China. \\
	$^{2}$University of Chinese Academy of Sciences, Beijing 100049, China. \\
	$^{3}$ Search Product Center, WeChat Search Application Department, Tencent, China.\\
}

\usepackage{times}
\usepackage{soul}
\usepackage{url}
\usepackage[hidelinks]{hyperref}
\usepackage[utf8]{inputenc}
\usepackage[small]{caption}
\usepackage{graphicx}
\usepackage{amsmath}
\usepackage{booktabs}
\usepackage{algorithm}
\usepackage{algorithmic}

\usepackage{times}
\usepackage{helvet}
\usepackage{courier}
\usepackage{amsfonts}

\usepackage{graphicx}
\usepackage{subfigure}
\usepackage{bm}
\usepackage{url}            
\usepackage{multirow}
\usepackage{rotating}
\usepackage{amsmath}
\usepackage{verbatim}
\usepackage{algorithm}
\usepackage{algorithmic}
\usepackage{comment}

\usepackage{enumitem}
\usepackage{amsmath,amssymb,amsthm}
\usepackage{lipsum}
\usepackage{float}

\usepackage[utf8]{inputenc} 

\title{Atom Responding Machine for Dialog Generation}

\begin{document}

	\maketitle
    
	\begin{abstract}
    Recently, improving the relevance and diversity of dialogue system has attracted wide attention.
    For a post $\bm{x}$, the corresponding response $\bm{y}$ is usually diverse in the real-world corpus, while the conventional encoder-decoder model tends to output the high-frequency (safe but trivial) responses and thus is difficult to handle the large number of responding styles.
    To address these issues, we propose the Atom Responding Machine (ARM), which is based on a proposed encoder-composer-decoder network trained by a teacher-student framework.
    To enrich the generated responses, ARM introduces a large number of molecule-mechanisms as various responding styles, which are conducted by taking different combinations from a few atom-mechanisms.
    In other words, even a little of atom-mechanisms can make a mickle of molecule-mechanisms.
    The experiments demonstrate diversity and quality of the responses generated by ARM. 
    We also present generating process to show underlying interpretability for the result.
\end{abstract}

\section{Introduction}

Recent years have witnessed the development of conversational models upon task-oriented system and free-chat system. Different from the traditional statistic models, the recent neural models directly learn the transduction from input post to output response in an end-to-end fashion, and outperforms the conventional models~\cite{Shang2015}.

Generally, the end-to-end model consists of two components: encoder and decoder. 
Inspired by the neural translation model ~\cite{Cho2014,Sutskever2014}, the encoder summarizes the user input $\bm{x}$ (post) into a context vector $\bm{c}$, and the decoder aims to decode $\bm{c}$ as the robot output $\bm{y}$ (response). To train this network, researchers usually directly maximizing the likelihood $p(\bm{y}|\bm{x})$ of post-response pairs.
So far, the encoder-decoder framework has been the state-of-the-art method for the dialog system. 
However, the relevance and diversity of the generated responses are still far from perfect.
Since some conventional models~\cite{Shang2015,Cho2014,Sutskever2014} are trained via directly maximizing the probability $p(\bm{y}|\bm{x})$, the generated responses are usually safe but trivial (e.g. ``It's OK'', ``Fine''). This may be due to the high frequency of these response types in the real-world corpus. 
However, the post-response style varies a lot from person to person which is quite natural in our daily life. Hence, generating the high-frequency response may damage the user experience of dialog system.
Meanwhile, good but rare responses often obtain low probability of $p(\bm{y}|\bm{x})$, it may be difficult for the conventional conversational model  to recall this type of response and thus reduce the response relevance.
Another underlying weakness of conventional encoder-decoder framework may be the lack of interpretability. Though many models can generate good responses, the responding process is still a black box which makes it difficult to explicitly control response content and style. This weakens the controllability of conventional models in real-world applications.

Recently, some models have been proposed to improve the relevance and especially the diversity of responses.
For example, Li et al.~\shortcite{Li2015} propose the Maximum Mutual Information (MMI) as  objective to improve response diversity. 
Specially, some researchers develop the mechanism-aware methods~\cite{Zhou2017MARM,zhou2018elastic}, which is based on the decomposition
$p(\bm{y}|\bm{x})\! = \!\sum_{i=1}^N p(\bm{y}|\bm{x}, m_i) p(m_i|\bm{x})$. In detail, mechanism-aware model assumes multiple corresponding mechanisms exist between post-response pairs (detailed in Section \ref{sec:pre}). By using  mechanism embedding $m_i$, the mechanism-aware methods learn the $1$-to-$n$ post-response relation (or discourse) in dialog corpus. Specially, with various mechanisms, the model may generate responses with different content and language styles for an input post, which improves the response diversity.
Furthermore, by explicitly selecting mechanisms, the mechanism-aware methods shows underlying ability of controlling the language styles of generated responses~\cite{zhou2018elastic}.
However, 
since real-world large corpus contains a huge amount of underlying pair-response relations,
the conventional mechanism-aware models may fail to model all the relations as the same number of latent vectors (mechanism embedding), and
may only cover a small part of them.

In our initial experiments, a certain post may correspond to various responses. Whereas, these responses may not be entirely different from each other. Some responses share similar linguistic structures. For example, for the post ``I have a exam tomorrow, and I am very nervous'', the response could be ``Relax, it would be fine'' or ``Don't worry, it would be fine'', where the two responses are actually quite similar.
Inspired by the observations, we assume that a post-response relation could be decomposed into a combination of sub-relations. The sub-relation is called as atom-mechanism. 
In this divide-and-conquer approach, a small number of atom-mechanisms can generate a huge number of combinations ($N$ atom-mechanisms to $\small(2^N-1\small)$ combination). This is what we say ``Even a little makes a mickle''.
For simplicity, the combination of atom-mechanisms is named as molecule-mechanism. 
As aforementioned, each molecule-mechanism corresponds to an implicit post-response relation. If two molecule-mechanisms consist of similar atom-mechanisms, they are likely to generate similar responses for a given post. We believe this divide-and-conquer approach may obtain a fine-grained control of responding, and strengthen the ability of modeling long-tail language styles in order to improve the response diversity and relevance.

Hence, we propose an Atom Responding Machine (ARM). In ARM, a post is responded by multiple molecule-mechanisms which consist of atom-mechanisms selected from global atom-mechanism set $\{m_i\}_{i=0}^N$.
Here, 
the atom-mechanisms are modeled as latent embedding vectors,
and a molecule-mechanism is modeled as a sequence of atom-mechanisms.
To generate the molecule-mechanism, the REINFORCE algorithm\cite{Williams1992Simple} is applied to train a proposed ``composer'' component which is a RNN decoder. For a post, based on the context vector $\bm{c}$, composer sequentially selects suitable atom-mechanisms as molecule-mechanisms for responding.
A molecule-mechanism corresponds to a composite function which transforms context vector $\bm{c}$ into mechanism-aware context vector $\widetilde{\bm{c}}$. At last, a decoder component receive $\widetilde{\bm{c}}$ and generate responses.

To this end, our contributions are summarized into three folds:
1) We propose an encoder-composer-decoder framework where suitable atom-mechanisms are selected to form molecule-mechanisms for response generation;
2) We propose a teacher-student framework to train the encoder-composer-decoder framework without supervised mechanism labels;
3) We empirically demonstrate that ARM generates more diverse and acceptable responses than previous methods, which might benefit from the fine-grained generation of molecule-mechanisms.
We also present generating process to show underlying interpretability for the result.


    \section{Preliminaries: Mechanism-aware Method}\label{sec:pre}

\subsection{Encoder-Decoder Framework}
Given an input post $\bm{x} = (x_1, x_2, \cdots , x_T)$ and an output response $\bm{y} = (y_1, y_2, \cdots , y_{T'})$, where $x_t$ and $y_t$ are the $t$-th word in post and response respectively.
The encoder-decoder framework~\cite{Cho2014,Sutskever2014} learns $p(\bm{y}|\bm{x})$ upon the training corpus $D = \{(\bm{x}, \bm{y})\}$ containing post-response pairs.

In detail, the conventional encoder-decoder model consists of two RNN components in succession: encoder and decoder. Encoder firstly summarizes the input sequence into a fixed-length context vector $\bm{c}$, and then decoder decodes $\bm{c}$ as the output sequence.
Both of encoder and decoder are RNNs.
They receive the previous hidden state $\bm{h}_{t-1}$ and current word embedding $\bm{x}_t$, and calculate $\bm{h}_{t} = f(\bm{h}_{t-1}, \bm{x}_{t})$, where $f$ is the activation function, e.g. LSTM~\cite{Hochreiter1997}, GRU~\cite{Cho2014} etc.
Due to space limitation, we omit the details of RNN computation.
\textbf{Mechanism-aware Method}

Some researchers develop the mechanism-aware methods~\cite{Zhou2017MARM,zhou2018elastic}.
To learn the $1$-to-$n$ post-response latent relation in dialog corpus, they utilize the following decomposition:
\begin{equation}
p(\bm{y}|\bm{x})=\sum_{i=1}^N p(\bm{y}|\bm{x}, m_i) p(m_i|\bm{x}),
\end{equation}
using responding mechanism $m_i$ generated from extra components.
Here, mechanism-aware method assumes $N$ latent mechanisms $\{m\}_{i=1}^N$ exists for a post-response pair.

In detail, as conventional encoder-decoder model, encoder of mechanism-aware method firstly summarizes the post as a context vector $\bm{c}$.
Then, a diverter, which is a softmax classifier, receives $\bm{c}$ as input, and output the distribution $p(m_i|\bm{x}) = \text{softmax}(\bm{m}_i \cdot \bm{c} + b_i)$ where $\bm{m}_i$ is mechanism embedding.
For MARM~\cite{Zhou2017MARM}, all mechanisms are used for generating response. For ERM~\cite{zhou2018elastic}, only a selected set of mechanisms are used. Note that for both of MARM and ERM, the number of generated responses is linear to the number of selected mechanisms.
While decoding with a selected $\bm{m}_i$, the decoder receives a mechanism-aware context vector $\widetilde{\bm{c}}_i=[\bm{m}_i; \bm{c}]$ (concatenation), and then decodes $\widetilde{\bm{c}}_i$ as the conventional decoder network, which recurrently updates the hidden state $\bm{s}_{t} = f(\bm{s}_{t-1}, y_{t-1}, \widetilde{\bm{c}}_i)$ and uses $\bm{s}_{t}$ to estimate the word probability $p(y_{t}|\bm{y}_{<t}, \bm{x})$. 

	\section{Atom Responding Machine}
Different from conventional mechanism-aware method,
ARM firstly selects groups of atom-mechanisms, then combines each group as a molecule-mechanism. At last, it uses each molecule-mechanism to generate a response.
Here, all the atom-mechanisms of the given corpus are assumed to be contained in the set $\{m_i\}_{i=0}^N$, where $m_0$ is an additional termination mechanism  (detailed later). 
Consistently with real-word application, the corpus for training is 
$\mathcal{D}=\{(\bm{x},\mathcal{Y}_{\bm{x}})\}$, 
where $\mathcal{Y}_{\bm{x}}$ is the set containing all the responses of $\bm{x}$.

\begin{figure}[t]\centering  

	\includegraphics[width=3in]{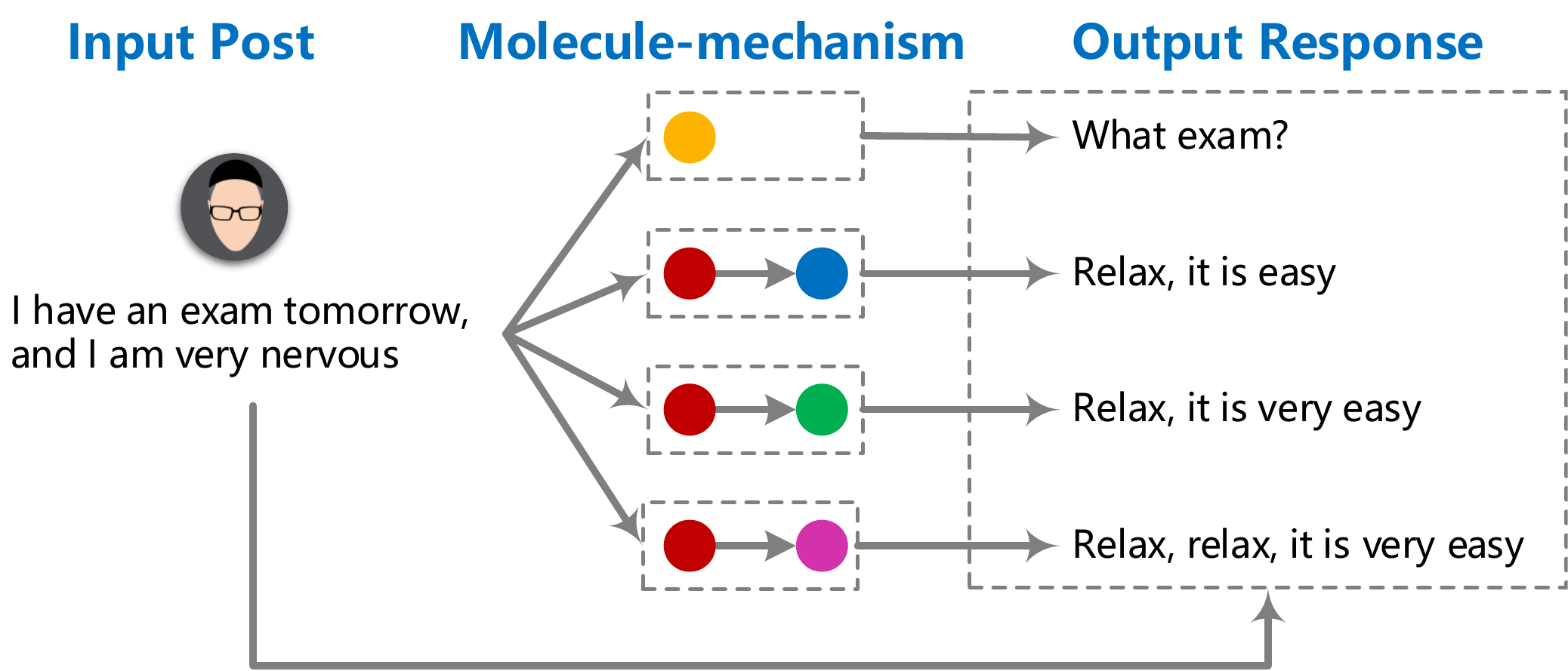}
	\caption{For an input post, multiple molecule-mechanisms could be utilized for responding.}
	\label{fig:atom_generate}
\end{figure}

Briefly, ARM contains three components: encoder, composer and decoder. The encoder summarizes the input post $\bm{x}$ as the context embedding $\bm{c}$, and then feeds $\bm{c}$ to the composer.
The composer is developed to select and combine atom-mechanisms as  molecule-mechanism using the fed context vector $\bm{c}$.
Note that in ARM, each atom-mechanism $m_i$ corresponds to a transformation function $g_i$. A molecule-mechanism $\mathcal{M}$ contains multiple atom-mechanisms and corresponds to a function $G_{\mathcal{M}} \equiv g_{i_{|\mathcal{M}|}} \circ g_{i_{|\mathcal{M}-1|}} \circ \cdots \circ g_{i_1}$ where ``$\circ$'' indicates function composition. Here, $\mathcal{M}$ denotes a molecule-mechanism. Then mechanism-aware context vector over $\mathcal{M}$ is defined as $\tilde{\bm{c}}_{\mathcal{M}}=G_{\mathcal{M}}(\bm{c})$.
At last, the decoder decodes $\widetilde{\bm{c}}_{\mathcal{M}}$ and output the final response.

Note that for real-world applications, it is unpractical to label all relations or discourses of post-response pairs as the supervised data for training. Hence, ARM has to infer the mechanism from post-response pair $(\bm{x},\bm{y})$. Here, we face two challenges: 
1) On one hand, the molecule-mechanism is determined by both post $\bm{x}$ and response $\bm{y}$. Hence, the composer is supposed to take both $\bm{x}$ and $\bm{y}$ as inputs.
2) On the other hand, the response $\bm{y}$ is only available in training. Thus, the composer has to generate all the corresponding molecule-mechanisms only according to $\bm{x}$ in testing.


To address the two challenges above, we divide ARM into two sub-networks: Teacher Network and Student Network. The two sub-networks obtain individual encoders, composers and decoders.
\begin{figure}[t]\centering
	\includegraphics[width=3in]{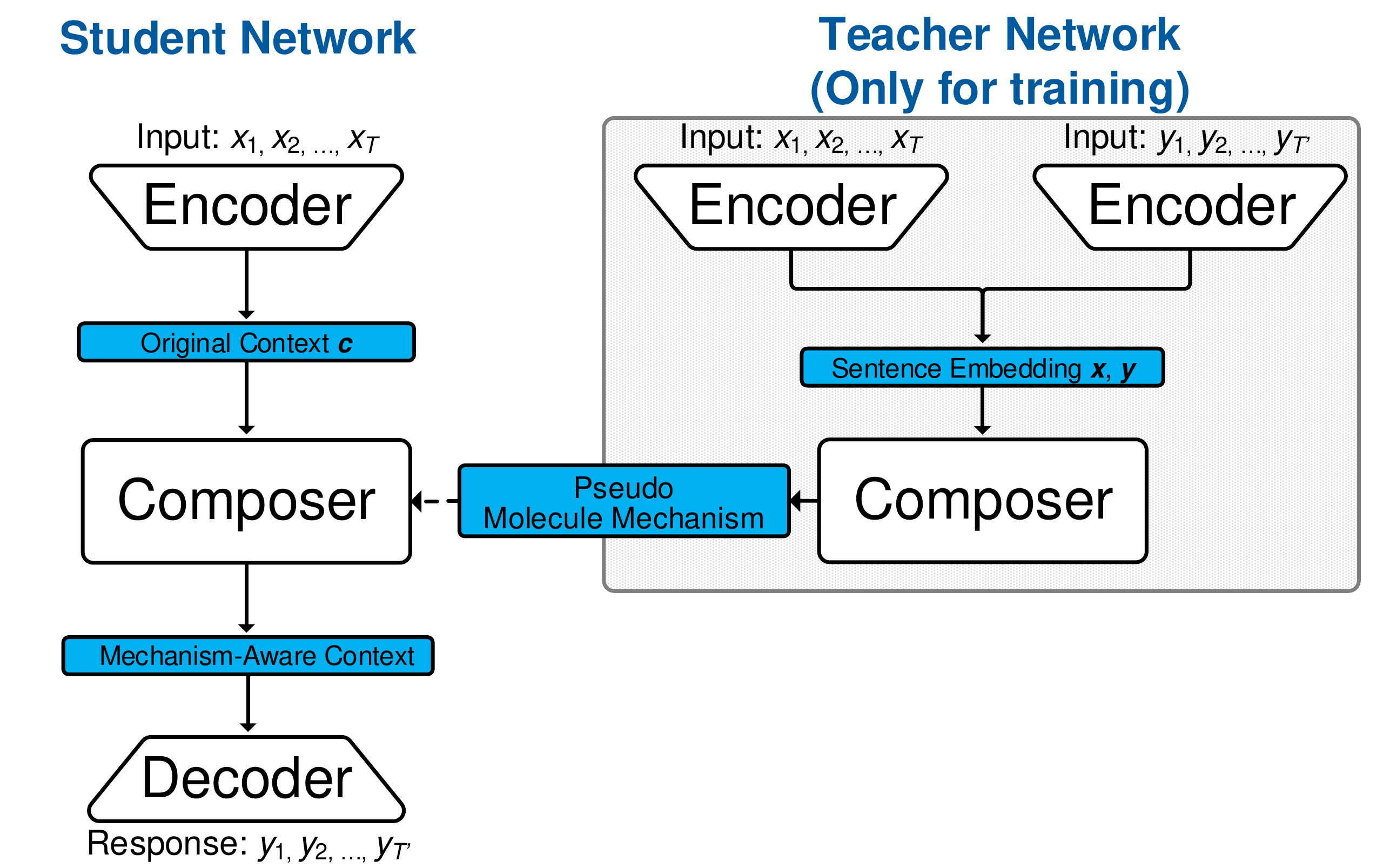}
	\caption{The structure of Atom Responding Machine.}
	\label{fig:arm_abstract_model}
\end{figure}
ARM's framework is shown in Fig.\ref{fig:arm_abstract_model}. Note that the encoders and decoders work as conventional RNN models, due to the space limitation, we omit their details. Next, we will introduce the Teacher Network and Student Network.

\textbf{Teacher Network:}
Teacher Network is used only in training. 
It contains two independent Teacher encoders which summarize $\bm{x}$ and $\bm{y}$ as vectors $\bm{c}_{\bm{x}}^\text{t}$ and $\bm{c}_{\bm{y}}^\text{t}$ respectively. Then, $\bm{c}_{\bm{x}}^\text{t}$ and $\bm{c}_{\bm{y}}^\text{t}$ are fed to Teacher composer
to generate a molecule-mechanism $\mathcal{M}^\text{t}=\mathcal{M}(\bm{x},\bm{y})$, which will be fed to Student Network as pseudo training data. Since the input of Teacher network contains response $\bm{y}$ which is  unavailable in testing, we call it as Teacher Network.

\textbf{Student Network:}
Student Network is used in both training and testing.
In detail, Student encoder firstly summarize the post $\bm{x}$ as vector $\bm{c}^\text{s}$. Then, Student composer receives $\bm{c}^\text{s}$ and outputs molecule-mechanism $\mathcal{M}^\text{s}$.
When training, the $\mathcal{M}^\text{t}$ generated by Teacher composer is the supervised data of Student composer.
When testing, Student composer generates multiple molecule-mechanisms $\mathcal{M}^\text{s}$ by maximizing $p(\mathcal{M}|\bm{x})$ via beam search.
At last, Using given $\mathcal{M}^\text{s}$, a Student decoder generates the responses $\bm{y}$ according to $\bm{c}^\text{s}$ and $\mathcal{M}^\text{s}$. 
Since the network generate molecule-mechanism $\mathcal{M}^\text{s}$ under the supervision of $\mathcal{M}^\text{t}$ of Teacher Network, we call this network as Student Network.

\subsection{Teacher Network}

It is intuitive that once $(\bm{x}, \bm{y})$ is given, their discourse or relation is then determined, and the molecule-mechanism is thus unique.
Hence, Teacher Network utilize both of $\bm{x}$ and $\bm{y}$ to generate molecule-mechanism $\mathcal{M}^\text{t}$.
Since $\bm{y}$ is only available at training period, thus the Teacher Network is only used in training and developed to guide Student Network learn how to generate suitable molecule-mechanisms. 
We then employ two independent Teacher encoders (RNN) and obtain context vectors $\bm{c}_{\bm{x}}^\text{t}$ and $\bm{c}_{\bm{y}}^\text{t}$ for post and response respectively (Fig.\ref{fig:arm_abstract_model}). Next, the vector $\bm{c}^\text{t}=[\bm{c}_{\bm{x}}^\text{t},\bm{c}_{\bm{y}}^\text{t}]$ is fed to Teacher composer.


Teacher composer works as a decoder, which utilizes $\bm{c}^\text{t}$ to generate an atom-mechanism sequence $\mathcal{M}^\text{t}$ (molecule-mechanism) ended with $\bm{m}_0$. Since there is no labeled molecule-mechanisms for supervision, we train the model in a sampling fashion as REINFORCE algorithm. In detail, the composer contains following components: 


\textbf{State:}
Here, we use $\mathcal{S}^{(k)}$ to denote the state at $k$-th time step. It stores partially generated molecule-mechanism sequence. 
At the beginning, $\mathcal{S}^{(0)}$ is initialized as am empty sequence. At $k$-th step, the model will sample and append an atom-mechanism $m^{(k)}$ to current state (controlled by policy).
Specifically, the state embedding is updated as:
\begin{equation}
\bm{h}^{(k)}=f(\bm{h}^{(k-1)}, \bm{m}^{(k)},\bm{c}^\text{t}),
\end{equation}
where $\bm{h}^{(k)}$ is the embedding of $\mathcal{S}^{(k)}$, $\bm{h}^{(0)}=\bm{0}$, $\bm{m}^{(k)}$ denotes the embedding of selected atom-mechanism $m^{(k)}$, and $f$ is a activation function (GRU\cite{Cho2014} is utilized in this work). According to the nature of RNN, $\bm{h}^{(k)}$ contains the information of partial molecule-mechanism to the $k$-th step.

\textbf{Action:}
We define $(N+1)$ actions $\{a_i\}_{i=0}^N$.
Specifically, $a_0$ indicates that the termination atom-mechanism $m_0$ is selected, which will terminate generating molecule-mechanism and output current state as generated molecule-mechanism $\mathcal{M}=\mathcal{S}^{(k)}$. Otherwise, if $a_i(i \neq 0)$ is triggered, $m_i$ is appended to the under-generating molecule-mechanism.

\textbf{Policy:}
Given the current state $\mathcal{S}^{(k)}$, the composer determines which action will be selected, and it is under the control of policy $\pi(\cdot|\bm{x},\bm{y},\mathcal{S}^{(<k)})=p(\bm{m}|\bm{h}^{(k)})$.
We calculate the policy as follows:
\begin{equation}
\begin{aligned}
\{a_i\}_{i=0}^N\sim\pi(\cdot|\bm{x},\bm{y},\mathcal{S}^{(<k)}) &\!=\!
\text{softmax}\big(W_{\pi}\bm{h}^{(k)}\big),
\end{aligned} 
\end{equation}
where $W_{\pi}\in\mathbb{R}^{(N+1)\times|h|}$ is a trainable parameter. After the policy $\pi(\cdot|\bm{h}^{(k)})$ is determined, we select action via sampling. 

\textbf{Reward:}
As aforementioned,
Teacher Network is developed to generate pseudo training data for Student Network.
Hence, we 
use $p(\bm{y}|\bm{x}, \mathcal{M}^\text{t})$ calculated by Student Network\footnote{Detailed at later section ``Calculating $p(\bm{y}|\bm{x},\mathcal{M})$''.} to define the reward.
In detail, for a set of molecule-mechanisms $\{\mathcal{M}^{\text{t}}_i\}$ generated by Teacher composer over $(\bm{x},\bm{y})$, the reward of the $i$-th molecule-mechanisms $\mathcal{M}^{\text{t}}_i$ is formulated as:
\begin{equation}\label{equ:teacher_reward}
r(\mathcal{M}^{\text{t}}_i)=
\frac{1}{|\bm{y}|}\big(\log p(\bm{y}|\bm{x},\mathcal{M}^{\text{t}}_i) - \min_{\mathcal{M'}\in\{\mathcal{M}^{\text{t}}_i\}}\log p(\bm{y}|\bm{x},\mathcal{M}')\big),
\end{equation}
where $|\bm{y}|$ denotes the length of response $\bm{y}$.


\subsection{Student Network}
Student Network learns to generate response only according to $\bm{x}$.
Since it is unpractical to label relation or discourse for all post-response pair in real-world corpus,
$\mathcal{M}^{\text{t}}$ is fed to Student Network as pseudo supervised data for mechanism in training. We then introduce how the Student composer and decoder estimate $p(\mathcal{M}|\bm{x})$ and $p(\bm{y}|\bm{x}, \mathcal{M})$ respectively.


\textbf{Calculating} $p(\mathcal{M}|\bm{x})$:
For a given $\bm{x}$, we denote $\{\mathcal{M}_i\}_{i=1}^{|\mathcal{Y}_{\bm{x}}|}$ as $\bm{x}$'s molecule-mechanism set, and define every response of $x$ corresponds to a unique molecule-mechanism.
Hence, the number of $x$'s responses\footnote{The duplicate responses of a post are removed.} equals to the number of $\bm{x}$'s molecule-mechanisms.

Student encoder firstly summarizes input $\bm{x}$ as a context embedding $\bm{c}^\text{s}$ using an independent encoder. Similar to Teacher composer, Student composer also works as a decoder, and receives context embedding $\bm{c}^\text{s}$  to generate molecule-mechanism $\mathcal{M}$.
In training,
since molecule-mechanism is a sequence, Student composer calculates the likelihood $p(\mathcal{M}|\bm{x})$.
Since conventional models train and generate via maximizing $p(\bm{y}|\bm{x})$, they tend to generate high-frequency responses (may be safe but vanilla).
To address this issue, for a given $\bm{x}$, all the molecule-mechanism probability $\{\mathcal{M}_i\}_{i=1}^{|\mathcal{Y}_{\bm{x}}|}$ is trained to be equal. Namely, the groundtruth probability $q(\mathcal{M}|\bm{x})=\frac{1}{|\mathcal{Y}_{\bm{x}}|}$.
To achieve this, using the molecule-mechanism $\mathcal{M}^\text{t}$ obtained from Teacher composer as pseudo supervised data, the KL-divergence $D_{\text{KL}} \big(p_{\theta_{\text{s}}}(\mathcal{M}^\text{t}|\bm{x}) \Arrowvert q(\mathcal{M}|\bm{x})  \big)$ 
is minimized as a part of objective in training, where $\theta_\text{s}$ is parameter of Student Network.
In testing, Student Network generate molecule-mechanisms via maximizing $p(\mathcal{M}|\bm{x})$.



\textbf{Calculating} $p(\bm{y}|\bm{x},\mathcal{M})$: 
To calculate $p(\bm{y}|\bm{x},\mathcal{M})$, we use a context vector $\bm{c}^\text{s}$ and a molecule-mechanism $\mathcal{M}$ to obtain a mechanism-aware context vector $\tilde{\bm{c}}$.
In training, $\mathcal{M}$ is $\mathcal{M}^\text{t}$ fed from Teacher Network. In testing, Student composer generates $\mathcal{M}$ by beam search.
Student decoder then utilizes $\tilde{\bm{c}}$ to calculate the $p(\bm{y}|\bm{x})$.

Next, we will detail how to obtain $\tilde{\bm{c}}$.
Each atom-mechanism $m_i$ corresponds to a transformation function $g_i$. For a molecule-mechanism $\mathcal{M}=(m_{i_1}, m_{i_2}, \cdots, m_{i_{|\mathcal{M}|}})$, it hence corresponds to a function $G_{\mathcal{M}}$ defined as follows:
\begin{equation}\label{equ:composition_function}
G_{\mathcal{M}} \equiv g_{i_{|\mathcal{M}|}} \circ g_{i_{|\mathcal{M}|-1}} \circ \cdots \circ g_{i_1},
\end{equation}
where ``$\circ$'' indicates function composition, and  $g_i(i=1,2,\cdots,N)$ is defined as
$
g_i(\bm{c}) \equiv f(\bm{c} + \bm{m}_i)
$,
where atom-mechanism embedding $\bm{m}_i$ is to be trained, $f$ is an user-defined function. ReLU is adopted as $f$ in this work~\cite{nair2010rectified}.
To this end, mechanism-aware context vector $\tilde{\bm{c}}=G_{\mathcal{M}}(\bm{c}^\text{s})$. Then, $\tilde{\bm{c}}$ is fed to Student decoder for calculating $p(\bm{y}|\bm{x},\mathcal{M})$.
\subsection{Training Details}

\textbf{Updating Teacher Network:} Teacher Network is trained by REINFORCE algorithm.
The basic idea is to maximize expected reward. In detail, for each $(\bm{x},\bm{y})$, Teacher Network first samples some molecule-mechanism $\mathcal{M}^\text{t}$.
These $\mathcal{M}^\text{t}$ are then fed to Student Network for calculating the likelihood $p(\bm{y}|\bm{x},\mathcal{M}^\text{t})$ as Equ.\eqref{equ:teacher_reward}.
Here, the sampled molecule-mechanisms are unique, as the responses to a post are different.



The parameters of Teacher Network are updated by maximizing expected reward, the gradient is calculated as:
\begin{equation}
\nabla_\theta J(\theta)= \mathbb{E}_{\mathcal{M}^\text{t}, a \sim \pi_\theta}[ \nabla_\theta \mbox{log}\pi_\theta\cdot(r(\mathcal{M}^\text{t})-b)],
\end{equation}
where the reward $r$ is considered as a constant while maximizing the expected reward, namely, the error back-propagation does not go through $r$. $\theta$ is model parameters.

\textbf{Updating Student Network:}
After Teacher Network generates molecule-mechanisms $\mathcal{M}^\text{t}$,
we randomly select one of them according to $p(\mathcal{M}^\text{t}|\bm{x},\bm{y})$ for training Student Network,
where $p(\mathcal{M}^\text{t}|\bm{x},\bm{y})$ indicates the probability of $\mathcal{M}^\text{t}$ generated by Teacher composer.
Then, we maximize the objective function by only adjusting parameters of Student Network:
\begin{equation}
	\max \sum_{(\bm{x},\bm{y})\in B}\frac{1}{|\bm{y}|}\!\log{p(\bm{y}|\bm{x}, \mathcal{M}^\text{t})}-D_{\text{KL}} \big(p(\mathcal{M}^\text{t}|\bm{x}) \Arrowvert q(\mathcal{M}|\bm{x})  \big),
\end{equation}
where $B$ indicates the current batch. Teacher Network and Student Network are updated alternately, namely, one network is fixed while updating the another.

\subsection{Generating Responses}
In this subsection, we will discuss how to generate response $\bm{y}$ for a given post $\bm{x}$. When generating responses in testing, only Student Network is utilized. The generating process obtains two steps: generating molecule-mechanism  and generating response .
\textbf{1) Generating Molecule-mechanism:}
Firstly, Student encoder calculates context vector $\bm{c}$. Then, based on $\bm{c}$, Student composer uses beam search to generate $L$ ($L$ is set manually) molecule-mechanisms via maximizing $p(\mathcal{M}|\bm{x})$. They are denoted as $\{\mathcal{M}_i\}_{i=1}^{L}$ in descending order of $p(\mathcal{M}|\bm{x})$. After that, these $L$ molecule-mechanisms are used to generate responses. 

\textbf{2) Generating Response:}
Next, for each of the generated molecule-mechanism $\mathcal{M}_i\in\{\mathcal{M}_i\}_{i=1}^{L}$, the Student Network calculates mechanism-aware context vector $\tilde{\bm{c}}_i=G_{\mathcal{M}_i}(\bm{c})$.
Student decoder then receives $\tilde{\bm{c}}_i$ and generates a response via maximizing $p(\bm{y}|\bm{x},\mathcal{M})$ using beam search. Finally, model outputs $L$ different responses.

	\begin{table*}[htbp]\small\setlength{\tabcolsep}{3.3pt}
	\centering
	\caption{The performance of each model. The ``Top-$k$'' denotes responses with top-$k$ probabilities in each group. Specially, ARM's responses are sorted in the descending order of $p(\mathcal{M}|\bm{x})$. }\label{table:rengong_ping_arm}
	\begin{tabular}{l|c|c|c|c|c||c|c|c||c||c|c}
		\hline
		\hline		
		\multirow{2}{*}{\textbf{Model}} &  \multicolumn{5}{c||}{\%\textbf{Acceptable}} &\multirow{2}{*}{\%\textbf{Bad} } & \multirow{2}{*}{\%\textbf{Normal}} & \multirow{2}{*}{\%\textbf{Good}} & \multirow{2}{*}{\textbf{BLEU-4}}& \multirow{2}{*}{\textbf{Top-5 Diversity}}& \multirow{2}{*}{\textbf{Total Diversity}}\\
		\cline{2-6}
		&Top-1&Top-2&Top-3&Top-4&Top-5&&&&&\\
		\hline
		\textsc{EncDec} & 31.78&	34.17&	37.11&	38.92&	40.47&	59.53&	35.93&	4.53& 8.78 & 0.3074 & -\\
		\textsc{Seq2Seq} & 45.00&	48.22&	48.56&	48.19&	48.40&	51.60&	40.80&	7.60& 12.45 & 0.2806 & -\\
		\textsc{Att} & 47.89&	49.89&	51.70&	53.00&	53.11&	46.88&	40.40&	12.71& 13.89 & 0.2760 & -\\
		NRM & 53.00&	54.39&	54.93&	55.42&	55.20&	44.80&	45.73&	9.47& 13.73 & 0.2434 & -\\
		\hline
		MMI-antiLM & 49.00 & 45.67 & 44.11 & 43.50 & 43.60 & 56.40 & 36.53 & 7.07& 8.56 & 0.2147 & -\\
		MMI-bidi & 58.67 & 58.33 & 55.89 & 54.67 & 54.60 & 45.40 & 45.07 & 9.53& 4.41 & 0.3060 & -\\
		\hline
		CVaR & 62.67 & 60.83 & 60.22 & 59.33 & 58.20 & 41.80 & 42.07 & 16.13& 2.94 & 0.5820 & -\\
		\hline
		CVAE & 65.00 & 66.33 & 65.78 & 65.75 & 65.07 & 33.93 & 41.43 & 24.73 & 5.37 & 0.6607 & -\\
		\hline
		MARM-4& 65.00 & 66.83 & 65.44 & 64.58 & 64.60 & 35.40 & 41.00 & 23.60& 11.56 & 0.5033 & -\\
		MARM-25&58.67&	54.83&	53.56&	53.00&	51.93&	48.07&	28.40&	23.53& 5.87 & 0.4280 & 0.2942\\
		\hline
		ERM&70.67&	72.83&	71.78&	71.75&	71.60&	29.85&	44.76&	25.39& 12.23 & 0.5493 & 0.5308\\
		\hline
		ARM & 74.33 &	74.83&	74.22&	73.92&	73.07&	27.87 & 33.77 & 38.37
		& 10.80 & 0.7307 & 0.7213\\
		\hline

	\end{tabular}

\end{table*}

\section{Experiment Process}
\subsection{Dataset Details}\label{sec:dataset}
All the models are trained on the dataset in \cite{zhou2018elastic} for experiments, which is collected from Tencent Weibo\footnote{http://t.qq.com/?lang=en\_US}. There are totally $815,852$ post-response pairs, among which $775,852$ are for training and $40,000$ for validating. $300$ posts are randomly sampled for testing which are not in the train and validation set. 

\subsection{Baseline Methods}
We use eleven conversation models for comparing:

1. A group of single-layer encoder-decoder models. \textsc{Seq2Seq}~\cite{Sutskever2014} and \textsc{EncDec}~\cite{Cho2014}: Two conventional models without auxiliary components. \textsc{Att}~\cite{Bahdanau2014Neural}: A model with attention mechanism. NRM~\cite{Shang2015}: A model with global and local scheme. Theses models generate five responses for each post.
    
2. MMI-bidi and MMI-antiLM~\cite{Li2015}: A single-layer encoder-decoder model with MMI penalty. We set $\lambda=0.5$ and $\gamma=1$. They generate five responses for each post.

3. CVaR: A single-layer encoder-decoder model. At each epoch, only responses with $\alpha=80\%$ smallest $p(\bm{y}|\bm{x})$ are used for training\cite{zhang2018tailored}. It generates five responses for each post.

4. CVAE: Zhao et al.~\shortcite{Zhao2017} proposed CVAE using conditional variational autoencoders. It generates 5 responses upon 5 respectively sampled $z$ using beam search.

5. MARM-4 and MARM-25~\cite{Zhou2017MARM}: 
A single-layer encoder-diver-decoder model. Here, MARM-4 use the default setting, where MARM-4 use four mechanisms for training and generate 5 responses using best two mechanisms for testing. MARM-25 uses 25 mechanisms for training, and generate 25 responses for testing (every mechanism will generate one response).
	
6. ERM-25 \cite{zhou2018elastic}: A single-layer elastic responding machine with 25 mechanisms (every mechanism will generate one response).
    
7. ARM: The proposed model. It totally contains $24$ atom-mechanism and a terminating  mechanism ($N=24$). For each testing post, Student Network generates ten molecule-mechanisms with highest $p(\mathcal{M}|\bm{x})$. Then, every molecule-mechanisms generates one response.

\subsection{Implement Details}
Vocabulary including 28,000 Chinese words are utilized in the experiments, which is segmented by the LTP tool \footnote{https://www.ltp-cloud.com/}. All the out-of-Vocabulary words are replaced with ``UNK''. 
For all the experimental models, the dimension of the word embedding is 128, the dimension of hidden state is 1024, and GRU~\cite{Cho2014} activation function is utilized. The parameters are sampled from a uniform distribution $[-0.01, 0.01]$. For training, ADADELTA~\cite{Zeiler2012ADADELTA} is used for optimization.
The training will be terminated if the error over the validation set increase for 7 consecutive epochs, and the model with the largest likelihood on validation set will be used for final comparison. When generating responses, we apply beam search with beam size $200$.

\subsection{Evaluation Measure}
\begin{table*}[t]
	\centering
	\caption{Response cases of experimental models.}
	\includegraphics[width=6.5in]{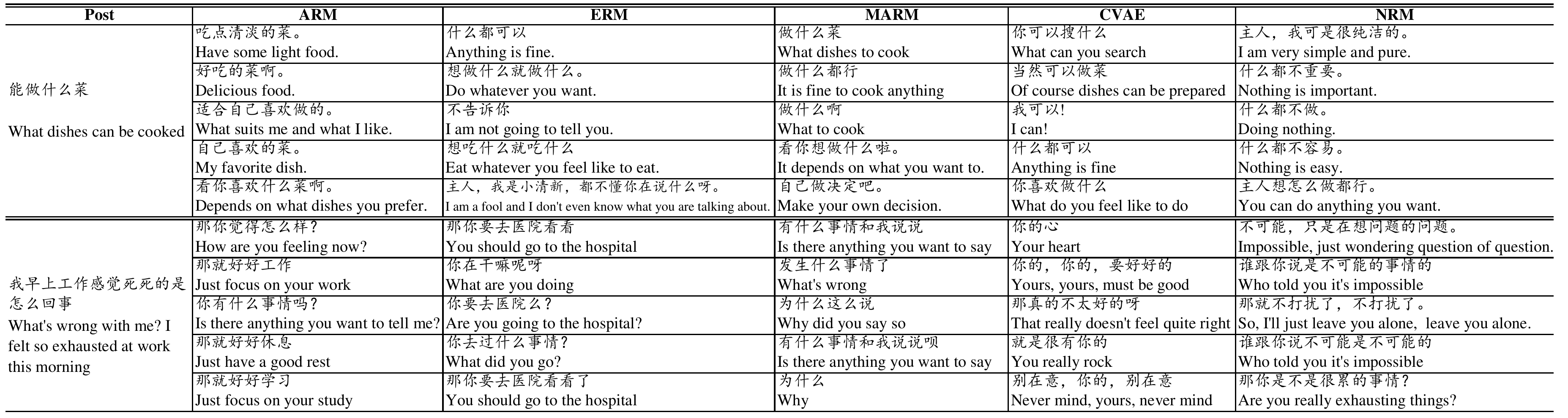}
	\label{fig:armcasestudy}
\end{table*}
We firstly utilize the three-level human judgment ~\cite{zhou2018elastic} to evaluate the responses.  Three labelers are invited to annotate the responses. \textbf{Bad}: The response is ungrammatical and irrelevant; \textbf{Normal}: The response is basically grammatical and relevant to the input post but vanilla and dull, e.g. ``Fine'' ``OK'' ``I don't know''; \textbf{Good}: The response is not only grammatical and relevant, but also meaningful and informative. Meanwhile, \textbf{Normal} and \textbf{Good} responses are also labeled as \textbf{Acceptable} responses.
Since three labelers are invited to label, the labeling agreement is evaluated by Fleiss' kappa~\cite{Fleiss1971Measuring} which is a measure of inter-rater consistency. Here, kappa value $\kappa=0.49$ (moderate agreement)\cite{landis1977measurement}.
Additionally, the diversity score is also reported.
For each post $\bm{x}$, the labelers annotate the number of different meanings among acceptable responses, namely $n$. Let $K$ denote the number of responses.
The diversity score for a post is $\frac{n}{K}$.
Then, a model’s diversity score is the average diversity score over the $300$ posts.
Additionally, ``Top-5 Diversity'' only statistics the top-5 responses, while ``Total Diversity'' statistics all responses.
To this end, BLEU is also reported~\cite{Papineni2002} to measure quality. However, BLEU is reported that may not be suitable for evaluation~\cite{Liu2016}. Hence, human judgment is considered as the main measure.

	\section{Experimental Results and Analysis}
\subsection{Experimental Results}

The results are summarized in the Table \ref{table:rengong_ping_arm}. Some experimental results are taken from \cite{zhou2018elastic}.

For the acceptable ratio and diversity, the best baseline method achieves 70.15\%  acceptable ratio, while ARM achieves 72.14\% with increase percentage of 2.84\%.
We further observe that, the improvement is mainly due to the ratio of Good responses of ARM is higher than the other baseline methods (38.37\% v.s. 25.39\%). This shows that ARM polishes some normal responses (high-frequency, trivial and boring) into more relevant responses which are high-quality.

Additionally, we observe ARM can generate some low-frequency and high-quality responses. This may be due to that ARM generates a response with the assistance of molecule-mechanism, and molecule-mechanism consists of atom mechanisms.
Note that ARM can fine-tune the molecule-mechanism by adding or removing some of its atom mechanisms, which makes the corresponding response fine-tuned as well.
We believe the flexible generation of molecule-mechanisms enables ARM to obtain fine-grained control of responding.
Hence, ARM may generate more low-frequency and high-quality responses.

We also observe the CVaR achieves 58.20\% diversity, while ARM achieves 72.13\% with increase percentage of 23.93\%.
We observe CVaR obtain unsatisfactory acceptable ratio.
In each epoch, since CVaR only uses responses with $80\%$ smallest likelihood for training, the response sets for training are different between epochs. Hence, responses may not be well fitted by CVaR and then ungrammatical.
For ARM, since ARM applies atom-level control for responding, we believe this improves diversity score of the model. 
Meanwhile, 
since the generating process of CVAE depends on randomly sampling latent varible $z$, it achieves satisfactory Top-5 diversity (66.07\%). However, the sampling may be unstable and let CVAE generate more ungrammatical responses which attack the acceptable ratio. To this end, ARM improves both the acceptable ratio and diversity in experiments. 

\begin{table}[t]

	\centering
	\caption{Response Examples of ARM, No.1}
	\resizebox{0.9\linewidth}{0.47\linewidth}{
		\includegraphics[width=2.9in]{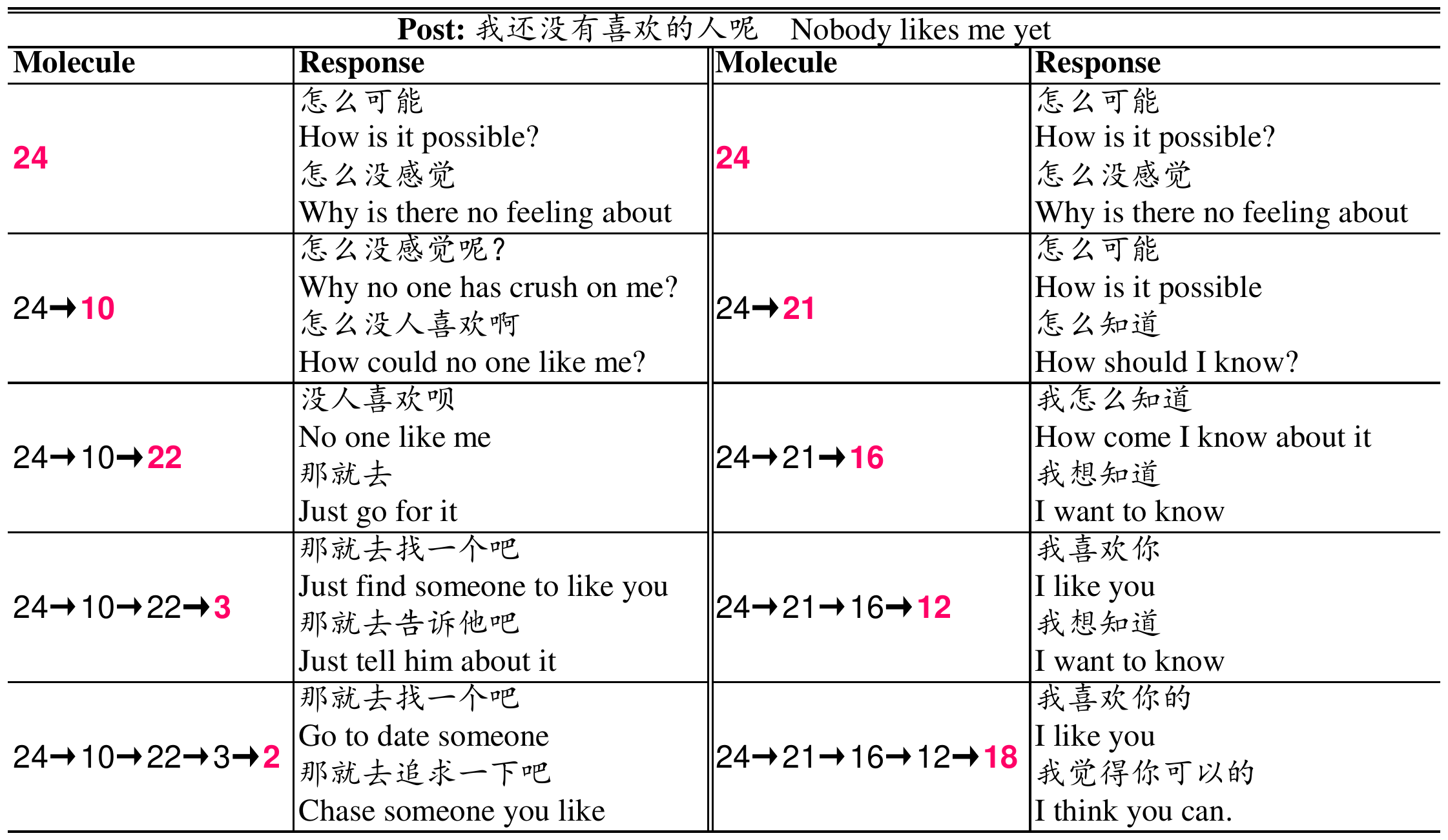}
	}
	\label{fig:arm_generating_process1}
	
	\caption{Response Examples of ARM, No.2}
	\resizebox{0.9\linewidth}{0.38\linewidth}{
		\includegraphics[width=2.9in]{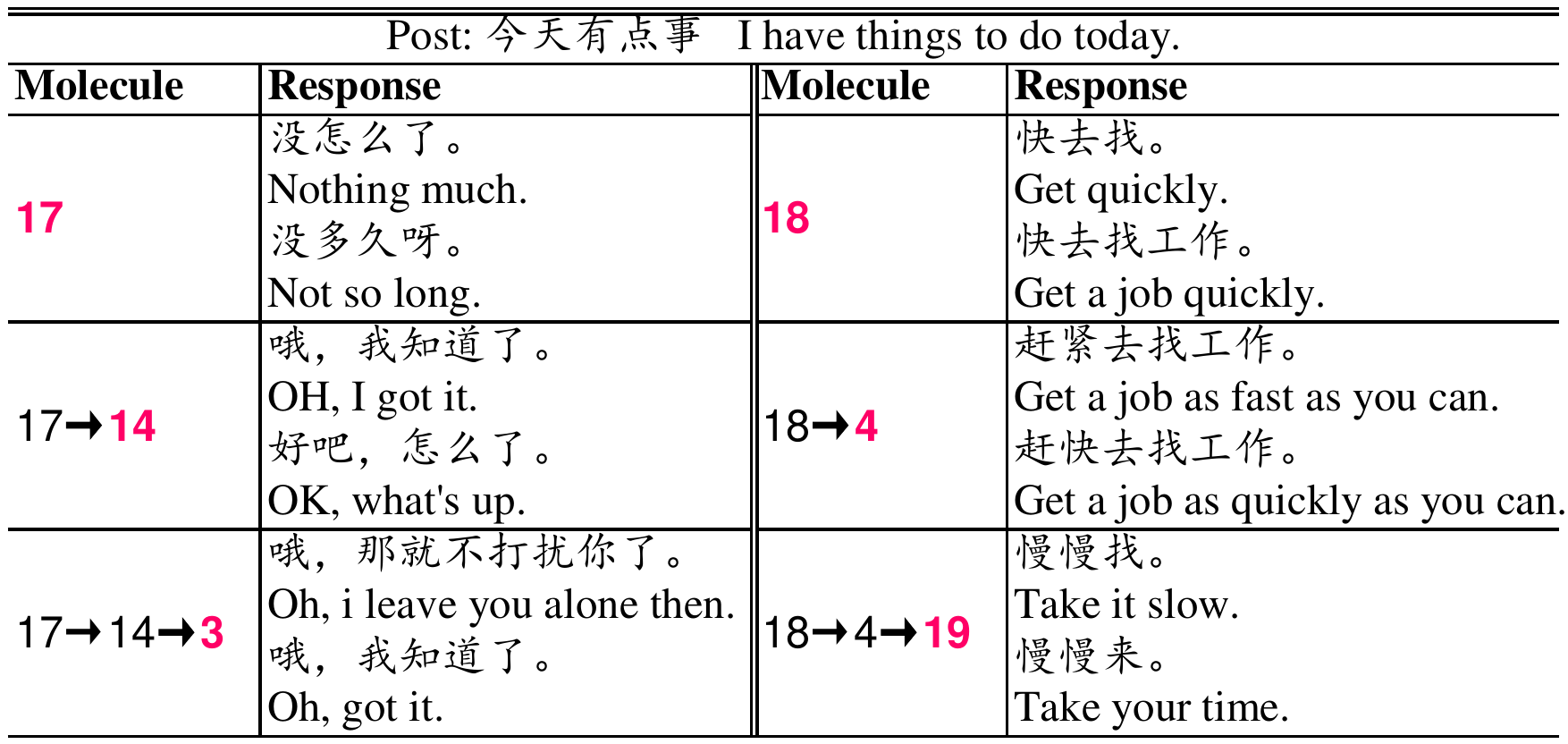}
	}
	\label{fig:arm_generating_process2}
\end{table}

	\begin{table*}[htbp]
	\centering
	\caption{Keywords of ARM's atom-mechanism (translated from Chinese). Here, results are sorted in descending order  of $p(w|m_i)$, and ``No.'' means the atom-mechanism index.}
	\includegraphics[width=5.5in]{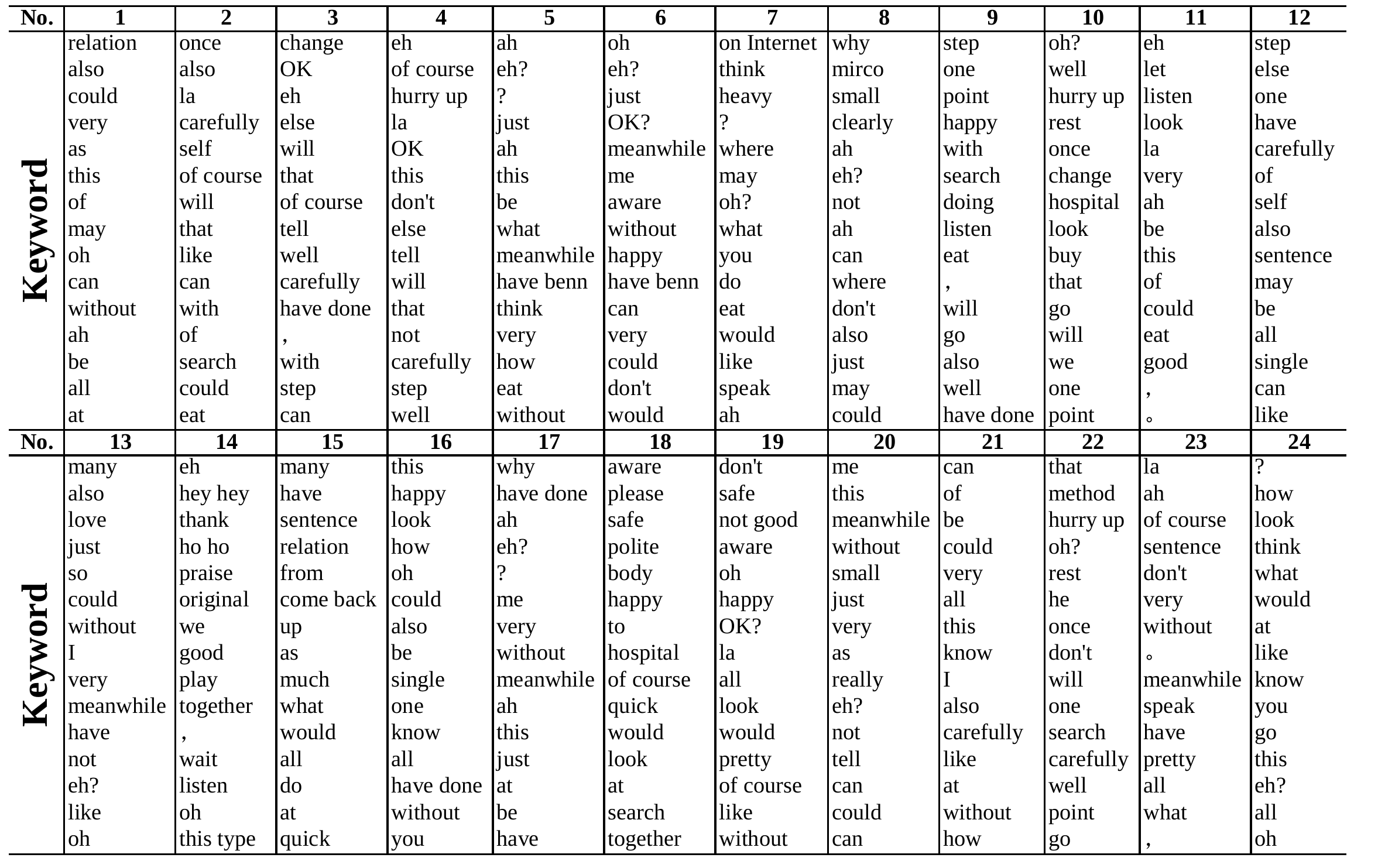}
	\label{fig:arm_keyword_EN}
\end{table*}
\subsection{Case Study}

The table \ref{fig:armcasestudy} shows some typical responses generated by ARM and other models. Each response is generated by a individual molecule-mechanism. 
Firstly, Table \ref{fig:armcasestudy} shows that ARM generates a more diverse set of responses compared to other models.
For example, for the post ``What dishes can be cooked'', ARM gives some typical responses, such as ``Have some light food.'', ``Delicious food.'' and ``What suits me and what I like.''. However, other models tend to repeat ``what to cook'', which leads to poor diversity.

Next, we further explore the relation between atom-mechanism and output response.
As aforementioned, a molecule-mechanism is a sequence of atom-mechanisms, and we therefore can utilize each prefix of a molecule-mechanism to generate responses respectively.
Table \ref{fig:arm_generating_process1} and \ref{fig:arm_generating_process2} show  response examples generated by prefixes of molecule-mechanisms. For example, ``24 $\rightarrow 10$'' denotes a molecule-mechanism consisting of the 24th and 10th atom-mechanism sequentially.
It shows that response is polished as atom-mechanism is appended to molecule-mechanism. For example, to the post ``Nobody likes me yet'', the generated response changes from ``How it is possible?'' to ``Chase someone you like''.
To explain how the atom-mechanism influences the generated response, we statistic the keywords (most frequent words) of each atom-mechanism. We discover that if a atom-mechanism $m$ is appended, the model tends to mix $m$'s keywords into response.
For example, since the top keywords of third atom-mechanism contains many exclamations like ``Just'', ``Oh'' and ``Eh'', molecule-mechanism ``$24\!\rightarrow\!10\!\rightarrow\!22\!\rightarrow\!\bm{3}$'' and ``$17\!\rightarrow\!14\!\rightarrow\!\bm{3}$'' mixed these exclamations into the response.

\subsection{Keyword of atom-mechanism}
Here, to explore how the atom-mechanism influences the generated response, we analyze the keywords of each atom-mechanism. In detail, for the $300$ testing posts, we generate $25$ molecule-mechanisms using beam search, and each molecule-mechanism generates $3$ responses. To remove the noise words, after some stop words are removed, only the words $w$ whose $p(w|m_i)>0.5$ are considered, where $p(w|m_i)$ denotes the probability that the word $w$ is generated if atom-mechanism $m_i$ is in the molecule-mechanism.

Table \ref{fig:arm_keyword_EN} show the keywords of each atom-mechanism. 
Firstly, we can observe that the keywords of atom-mechanisms are visibly different. Hence, for molecule-mechanism consisting of different atom-mechanisms, the generated responses tend to be different.

We also observe that the keyword genre of an atom-mechanism may influence the genre of generated response. For example, for the atom-mechanism No.5, No.17 and No.24 whose keyword lists contain many interrogative words, we observe that if a molecule-mechanism prefers such type of atom-mechanism, the generated response also tends to be interrogative. Hence, we conjecture that the response language style may be explicitly controlled by selecting suitable atom-mechanisms according to their keywords, and we will explore it in the future.	

Hence, we believe ARM may be more interpretable. 
It may be a promising way to control the language style and genre via selecting atom-mechanism in specific language style or sentiment. We will explore training the atom-mechanism using auxiliary sentiment information and explicitly control the response sentiment in the future work.
\section{Related Work}\label{sec:relatedWork}
\textbf{End-to-end Neural Network.} 
The generative models which is based on encoder-decoder framework has achieved great success in both statistical machine translation (SMT) and conversational models. Specially, many researchers propose models for single-round conversation~\cite{Sutskever2014,Cho2014,Bahdanau2014Neural,Shang2015,Wang2016}. Here, the diversity problem is ignored and the model tends the generate the high-frequency response which is safe but vanilla.
Recently, researchers begun to investigate models for multiple-round conversation. Serban et al.~\shortcite{Serban2015} built a generative hierarchical neural network. A related model proposed by Sordoni et al.~\shortcite{Sordoni2015} applied a hierarchical recurrent encoder-decoder model for query suggestion. The basic idea for multiple-round conversation is to extend the context generation from the immediate previous sentence to several previous ones.
Serban et al.\shortcite{Serban2017b} then proposed a hierarchical RNN model modeling complex dependencies between sub-sequences. Here, ARM can be applied to the multi-round conversation and improve the responding performance of utterance in another vertical direction.

\textbf{Improving Response Diversity.}
Some other models are proposed to tackle response diversity problem.
Li et al.~\shortcite{Li2015} proposed the Maximum Mutual Information (MMI) as the objective to improve the diversity. Our experiments show that it decreases the acceptable ratio.
Li et al.~\shortcite{li2016deep} proposed a reinforcement dialog generation model using MMI to generate informative, coherent, and easy-to-answer responses. Note that its reinforcement module is not designed for controlling the responding mechanisms.
Zhou et al.~\shortcite{Zhou2017MARM} applied a quantitative study on the diversity problem. They then proposed MARM to generate diverse responses with different mechanisms. However, its mechanism number for responding needs to be handcrafted and might not be satisfactory for every post.
Later, Zhou et al.~\shortcite{zhou2018elastic} applied reinforcement framework ERM, which generate different mechanisms for a given post. However, the mechanism number may be too large for large-scale data where the model may fail.
Zhao et al.~\shortcite{Zhao2017} proposed CVAE and kgCVAE using conditional variational autoencoders. Zhang et al.~\shortcite{zhang2018tailored} proposed a CVaR-based method which focusing on improving responses with low likelihood. However, these methods may be difficult to explicitly control responding mechanism, or may need extra feature engineering of discourse and dialog act during training (e.g. kgCVAE) which limits its applications in real-world corpus.

\section{Conclusion}\label{sec:con}
In this study, we proposed an Atom Responding Machine (ARM), which utilizes the molecule-mechanisms to model latent pair-response relation as generating mechanism. Molecule-mechanism consists of atom mechanisms, which enables the model to control the responding mechanism at microscale. Meanwhile, ARM can construct various molecule mechanisms with a small number of atom-mechanism. The experiments demonstrate ARM could generate more acceptable and diverse response compared to some strong baselines.

ARM is a promising solution to control the language style at microscale via controlling atom-mechanism selecting. Different from conventional RNN models, we believe ARM is a half white-box system. Here, we sequentially compose the molecule mechanism. In the future, composing the molecule mechanism using more complex structure (e.g. tree, graph) may further improve the model performance.  Meanwhile, we conjecture that the response language style may be explicitly controlled by selecting suitable atom-mechanisms based on keywords, and we will explore it in the future.




	\bibliographystyle{named}
	\small
	\bibliography{docs}
	
\end{document}